\PassOptionsToPackage{unicode}{hyperref}
\PassOptionsToPackage{hyphens}{url}

\documentclass{article}
\usepackage[final,nonatbib]{neurips_2024}

\makeatletter
\renewcommand{\@noticestring}{}
\makeatother

\usepackage[utf8]{inputenc}
\usepackage[T1]{fontenc}
\usepackage{hyperref}
\usepackage{url}
\usepackage{booktabs}
\usepackage{amsfonts}
\usepackage{nicefrac}
\usepackage{microtype}

\usepackage{xcolor,colortbl}
\usepackage{amsmath,amssymb,amsthm}
\usepackage{graphicx}
\usepackage{subcaption}
\usepackage{enumitem}
\usepackage{multirow}
\usepackage{float}
\usepackage{array}
\usepackage{rotating}

\graphicspath{{figures/}}

\usepackage{caption}
\title{Reinforcement Learning in Super Mario Bros:\\
  Curriculum, Pedagogy, and Optimal Level Design in World 1-1}

\author{%
  Jesse Ponnock \\
  Johns Hopkins University \\
  \texttt{jponnoc1@jhu.edu}
  \And
  Lucas Ho \\
  Johns Hopkins University \\
  \texttt{lho7@jhu.edu}
}

\begin{document}
\maketitle

\begin{abstract}
World 1-1 of Super Mario Bros is widely celebrated as a masterclass
in game design: its progressive structure is credited with teaching
players core mechanics through the level itself. We ask whether that
structure is empirically measurable using reinforcement learning.
We implement World 1-1 from scratch as a fully discrete environment
and compare four algorithms---Q-Learning, SARSA, Monte Carlo, and
Deep Q-Network (DQN)---across three progressively complex versions
of the same level. Monte Carlo emerges as the strongest agent
($94.9\% \pm 1.5\%$ win rate), outperforming DQN ($76.4\% \pm 3.4\%$)
by learning to maximize intermediate rewards along winning paths rather
than taking the most direct route. We then use Monte Carlo in a
curriculum experiment permuting World 1-1's six canonical segments
across twelve conditions. Canonical ordering converges fastest,
achieves the highest learning efficiency, and is the only condition
with zero catastrophic failures; no random permutation matches all
three criteria simultaneously. These results provide, to the best of
our knowledge, the first empirical validation that World 1-1's
canonical design encodes genuine pedagogical structure: one that
measurably accelerates learning and cannot be replicated by chance.
\end{abstract}

\section{Introduction}

Super Mario Bros World 1-1 is widely cited as a masterclass
in game design. The level opens with open ground and a single
enemy, introduces pipes and gaps mid-level, and reserves enemy
clusters and staircases for the final approach. This progressive
structure is widely credited with teaching core mechanics
through safe, low-stakes encounters before demanding their
integration \cite{dahlskog2012,cao2022,robinson2015}. Whether that structure is genuinely pedagogical, or simply a
compelling narrative that has grown around a well-made game, has
never been formally tested. Here, we use reinforcement learning as
an empirical lens to find out.

We implement World 1-1 as a fully discrete environment: Mario's
position is represented as tile coordinates, movement as tile steps,
and enemy behavior follows deterministic rules. This is a faithful
simplification of a game whose grid-based design makes discretization
natural. This makes it an ideal domain for comparing tabular and
function-approximation methods, as tabular Q-tables are \emph{exact}
on a discrete state space; they do not approximate the value function,
they record it. Any advantage DQN holds is therefore purely from
generalization across states, not from necessity.

The standard narrative in reinforcement learning is that DQN
scales to problems tabular methods cannot \cite{mnih2015}.
As we layer additional mechanics into our implementation (from
a static level to one with interactive objects to one with
enemies), the state space grows and function approximation
becomes an increasingly reasonable tool. If DQN's generalization
provides a meaningful benefit, it should become apparent as
complexity increases. We therefore test each algorithm across
three versions of the level of increasing complexity to observe
whether and when that crossover occurs.

\subsection*{Research Questions}

\begin{enumerate}
  \item Which algorithm learns World 1-1 most effectively, and
        does the answer depend on level complexity?
  \item Does World 1-1's canonical segment ordering accelerate
        RL learning compared to randomized equivalents?
\end{enumerate}

\subsection*{Contributions}

\begin{itemize}
  \item A custom Python implementation of World 1-1 with
        NES-faithful physics and three levels of complexity,
        built without any game or RL environment library.
  \item An algorithm comparison across 60 training runs
        (4~agents $\times$ 3~levels $\times$ 5~seeds).
  \item A curriculum learning experiment: 60 additional runs
        across 12~conditions sampling the permutation space of a
        6-segment level partition.
\end{itemize}

\section{Methods}

\subsection{Environment Design}

\subsubsection{Level Representation}

The environment is a $212 \times 14$ tile grid representing
World 1-1. We implement three progressively complex versions:

\begin{itemize}
  \item \textbf{v1 (static)}: Ground, pipes, gaps, and the
        end-of-level staircase. No interactive objects.
  \item \textbf{v2 (+blocks)}: Adds breakable brick blocks and
        hittable question blocks at their NES-accurate positions.
  \item \textbf{v3 (+enemies)}: Adds 17 enemies (16~Goombas
        and 1~Koopa, treated uniformly) with NES-accurate speed
        ($\frac{1}{3}$ Mario's speed per tile step), gravity,
        cascade group activation, and a stomp kill mechanic.
\end{itemize}

\begin{figure}[htbp]
  \centering
  \includegraphics[width=\textwidth]{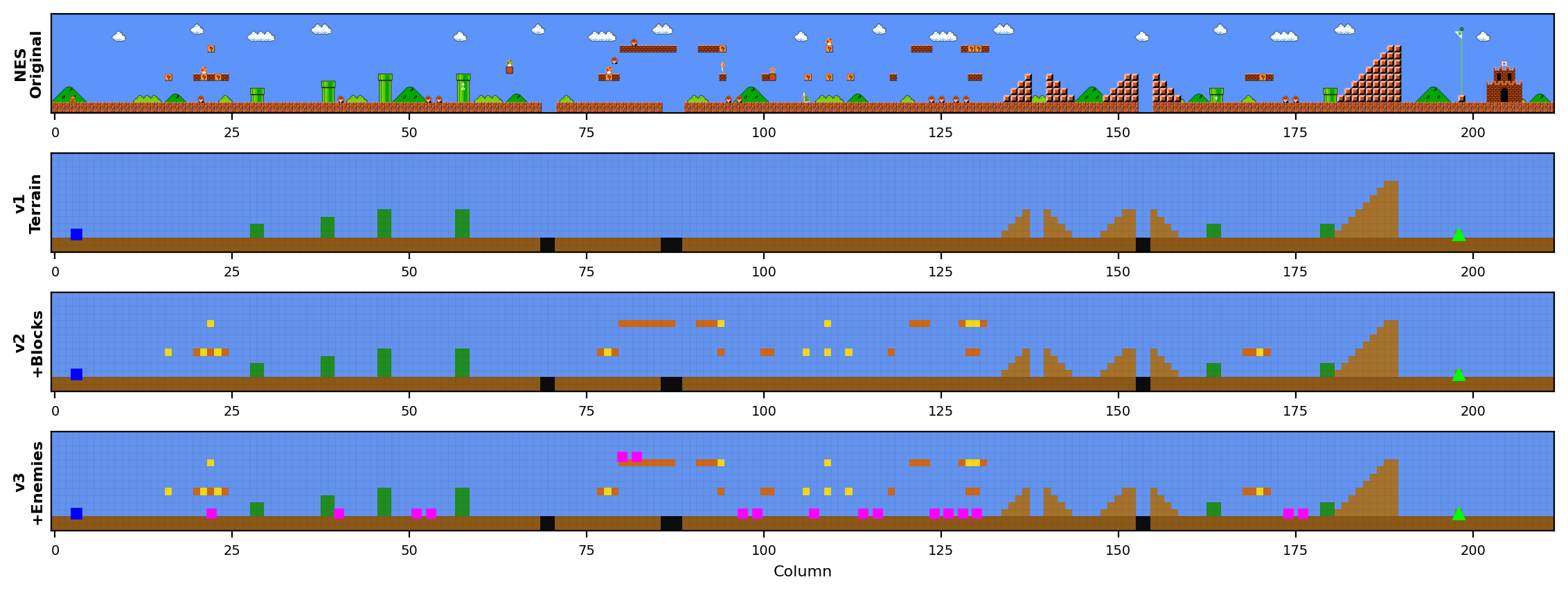}
  \caption{World 1-1 from NES original (top) to our three implemented
    versions. The tile grid reproduces all structural features of the
    original: ground terrain, gaps, pipes, and the end-of-level
    staircase (v1); breakable brick and question blocks at NES-accurate
    positions (v2); and 17 enemies with NES-faithful movement (v3).}
  \label{fig:level_comparison}
\end{figure}

Figure~\ref{fig:level_comparison} shows the NES original alongside
our tile-grid discretization and all three level versions.

\subsubsection{State Space}

The state representation grows with level complexity:

\textbf{v1:} $s = (\textit{col},\, \textit{row},\, \textit{jump\_phase})$.
Column indexes Mario's tile (0--211), row indexes his tile
row (0--13, 0 at top), and jump phase encodes position within
the fixed arc ($0 = $ grounded, $1$--$6 = $ in-air frame).

\textbf{v2:} $s = (\textit{col},\, \textit{row},\, \textit{jump\_phase},\, \textit{viewport})$,
where \textit{viewport} is a 56-value tuple encoding tile codes
at the two block rows (4 and 8) within a 28-column window
centered on Mario. The window width matches the NES screen width,
giving the agent exactly the information a human player sees.
This lets the agent learn \emph{blocks nearby $\Rightarrow$ jump
is worthwhile} rather than memorizing specific columns.

\textbf{v3:} The viewport extends to 112 values, adding two
enemy rows (rows 7 and 11, where Goombas walk) encoded as
\texttt{GOOMBA=7} or \texttt{EMPTY=0}. Block and enemy rows
occupy distinct positions in the tuple, so the agent can
unambiguously distinguish terrain from enemies.
Total state tuple length: 115.

All three Q-tables are implemented as \texttt{defaultdict} of
\texttt{defaultdict(float)}, so only visited states consume
memory regardless of the theoretical state-space size.

\subsubsection{Action Space}

Three discrete actions are available at every timestep:

\begin{table}[htbp]
  \centering
  \caption{Action space. Jump is grounded only; horizontal movement remains available while airborne.}
  \label{tab:actions}
  \begin{tabular}{cl}
    \toprule
    Code & Action \\
    \midrule
    0 & Stand still (grounded or airborne) \\
    1 & Move right (grounded or airborne) \\
    2 & Jump (grounded only; ignored if airborne) \\
    \bottomrule
  \end{tabular}
\end{table}

Left movement is omitted. World 1-1 is a strictly left-to-right
level with the goal always to the right; adding left movement
would expand the state space and exploration burden substantially
for an action no forward-progressing policy needs. The win reward
($+100$, described in Section~\ref{sec:reward}) dominates any
marginal gain from backtracking to collect additional blocks or
enemies, making left movement strategically redundant.

The jump arc is fixed and deterministic: once initiated, the
six-phase vertical trajectory $(-2, -3, -4, -4, -3, -2)$ tile
offsets cannot be altered. Horizontal movement remains
agent-controlled during the arc, giving variable jump distance
without a separate ``jump right'' action.

\subsubsection{Reward Function}
\label{sec:reward}

\begin{table}[htbp]
  \centering
  \caption{Reward function. ``v2+'' = applies to v2 and v3; ``all'' = all three versions.}
  \label{tab:rewards}
  \begin{tabular}{lcc}
    \toprule
    Event & Reward & Level \\
    \midrule
    Reach flagpole (win) & $+100$ & all \\
    Death (gap or enemy) & $-50$ & all \\
    Rightward step & $+0.5$ & all \\
    Stand still & $-0.1$ & all \\
    Break brick block & $+1$ & v2+ \\
    Hit question block & $+3$ & v2+ \\
    Defeat Goomba (stomp) & $+5$ & v3 \\
    \bottomrule
  \end{tabular}
\end{table}

The directional reward $+0.5$/step is calibrated so that the
total reward for traversing the level (${\approx}99 \times 0.5
= 49.5$) is co-primary with the win bonus ($+100$). This
prevents the degenerate policy of dying immediately (negative
reward) or ignoring the win condition in favor of farming
directional reward. The calibration follows the x-position
delta reward used in Kautenja~\cite{kautenja2018}.

\subsection{Algorithms}

All four algorithms use $\epsilon$-greedy exploration with a
shared decay schedule: $\epsilon$ starts at 1.0, decays
multiplicatively by 0.999 per episode, and floors at 0.01 for
tabular methods and 0.05 for DQN; the higher floor for DQN
follows \cite{mnih2015} and ensures the replay buffer retains
a diverse mix of transitions rather than collapsing to
near-identical on-policy samples as exploration decays.
Discount factor $\gamma = 0.99$ throughout.

\subsubsection{Q-Learning}

Off-policy temporal-difference learning \cite{watkins1992}.
At each step the update is:
\begin{equation}
Q(s,a) \leftarrow Q(s,a) + \alpha\bigl[r + \gamma \max_{a'}Q(s',a') - Q(s,a)\bigr]
\label{eq:qlearning}
\end{equation}
with learning rate $\alpha = 0.1$.

\subsubsection{SARSA}

On-policy temporal-difference learning \cite{rummery1994}.
Like Q-Learning but bootstraps from the \emph{next chosen
action} $a'$ rather than the greedy action:
\begin{equation}
Q(s,a) \leftarrow Q(s,a) + \alpha\bigl[r + \gamma Q(s',a') - Q(s,a)\bigr]
\label{eq:sarsa}
\end{equation}

\subsubsection{Monte Carlo}

First-visit, sample-mean Monte Carlo \cite{sutton2018}.
The agent buffers the full episode trajectory and updates at
episode end. For each first-visit $(s_t, a_t)$:
\begin{equation}
Q(s_t, a_t) \leftarrow Q(s_t, a_t) + \frac{1}{N(s_t,a_t)}\bigl[G_t - Q(s_t,a_t)\bigr]
\label{eq:mc}
\end{equation}
where $G_t = \sum_{k=0}^{T-t-1} \gamma^k r_{t+k+1}$ is the
discounted return from step $t$ and $N(s_t,a_t)$ is the visit
count. Unlike TD methods, MC does not update until an episode
terminates; early deaths produce short trajectories with
little signal about later states in the level.

\subsubsection{Deep Q-Network (DQN)}

A PyTorch multi-layer perceptron replacing the Q-table \cite{mnih2015}.
Architecture: input $\to$ 128 $\to$ 128 $\to$ 3 (ReLU
activations). The network is built lazily on the first observed
state, making it compatible with all three level versions
automatically. Key hyperparameters:

\begin{itemize}
  \item Experience replay buffer: 50,000 transitions
  \item Target network sync: every 500 gradient steps
  \item Training frequency: every 4 environment steps
  \item Learning rate: $10^{-4}$
  \item Batch size: 64
\end{itemize}

Hyperparameters were arrived at through empirical tuning on v1:
a 10k replay buffer caused replay buffer imbalance as death
transitions flooded and overwrote successful experiences; a
learning rate of $10^{-3}$ caused destabilizing weight updates;
a target-sync period of 100 steps caused the agent to chase a
rapidly moving regression target. Each modification follows
standard DQN practice \cite{mnih2015}.

\subsection{Experimental Setup}

\subsubsection{Training Protocol}

Each agent-level combination was trained for 10,000 episodes with five independent random seeds (42--46), extended to ten seeds (42--46, 54--58) when testing the MC agent's performance on canonical and reversed conditions in 3.2 for improved statistical power. Results are reported as mean $\pm$ standard deviation across seeds. 

\subsubsection{Statistical Analysis}

Pairwise comparisons use Welch's t-test with Cohen's \textit{d}. The DQN null curriculum effect uses a one-way ANOVA with $\eta^2$. 

\subsubsection{Evaluation Metrics}

\begin{itemize}
  \item \textbf{Win rate}: fraction of episodes completed
        (flagpole reached) over the final 500 episodes
        (episodes 9,501--10,000).
  \item \textbf{Average return}: mean undiscounted episode
        return over the final 500 episodes.
  \item \textbf{Convergence speed}: episodes to first reach a
        smoothed (100-episode rolling window) win rate of 50\%
        and 80\%, respectively.
  \item \textbf{Interaction metrics} (v2/v3 only): mean bricks
        broken, question blocks hit, and enemies defeated per
        episode over the final 500 episodes.
\end{itemize}

\subsubsection{Curriculum Conditions}

\begin{figure}[htbp]
  \centering
  \includegraphics[width=0.8\textwidth]{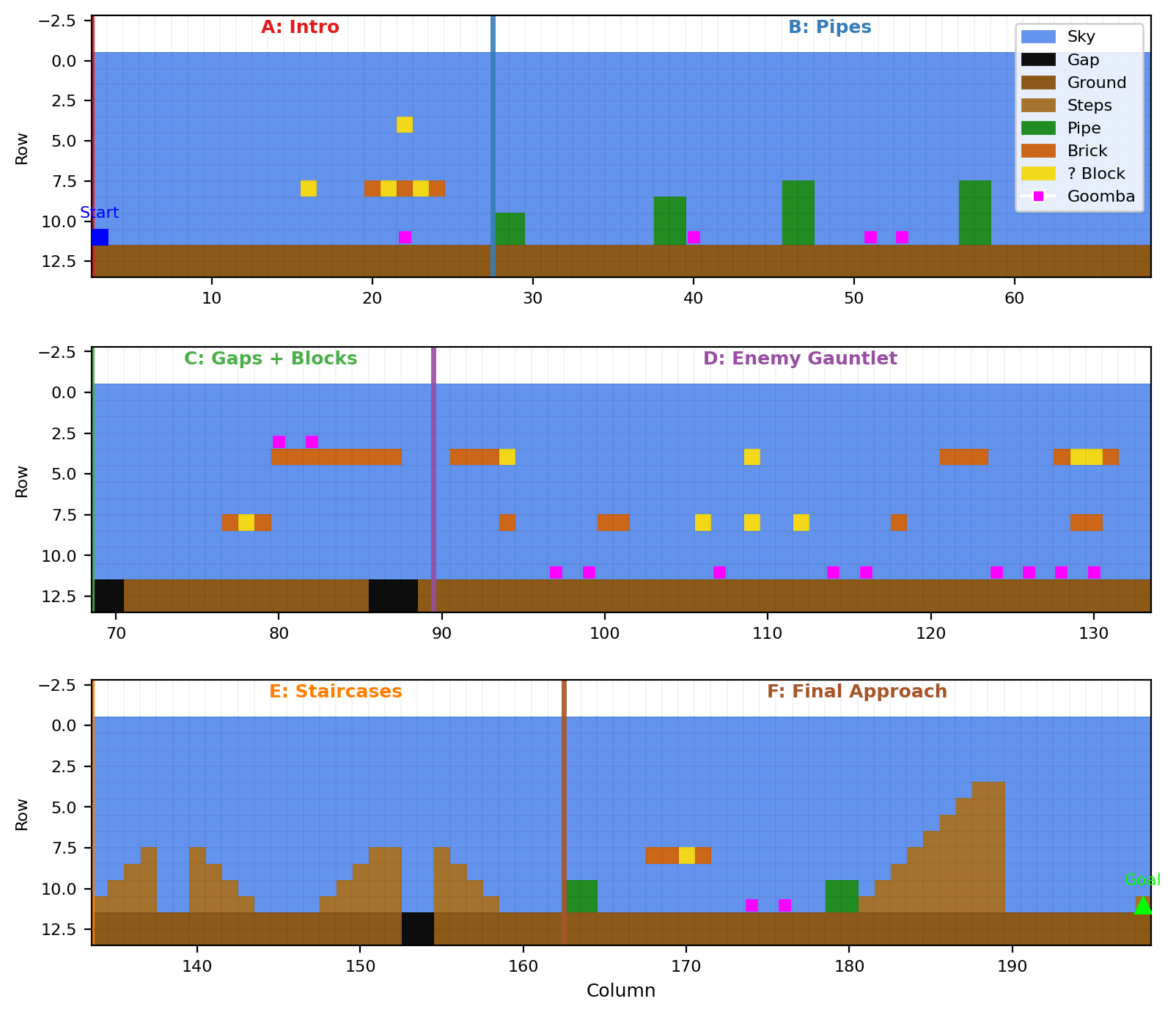}
  \caption{World 1-1 level map (v3) partitioned into six curriculum
    segments (A--F). All terrain features and enemy placements from
    the NES original are preserved. In the curriculum experiment,
    all six segments appear in every condition; only their sequence
    differs.}
  \label{fig:level_map}
\end{figure}

The level is partitioned into six segments (A--F) shown in
Figure~\ref{fig:level_map}. All six segments appear in every
condition; only their order differs. Content (enemy count, gap
count, block count) is identical across all orderings.

Twelve conditions were tested: the canonical ordering
(A$\to$B$\to$C$\to$D$\to$E$\to$F), the reversed ordering
(F$\to$E$\to$D$\to$C$\to$B$\to$A), and ten random permutations.
Permutation seeds are fixed independently from training seeds
(ordering seeds 200--209; training seeds 42--46), so map index
$i$ always produces the same segment order. The flagpole is
always fixed at column 198 regardless of segment order. A 3-tile
landing zone at Mario's spawn is always cleared to prevent spawn
collisions when segments that begin with elevated terrain are
placed first.

The Monte Carlo agent was selected for the curriculum experiment
because (1) it is the strongest v3 agent and (2) its
episode-level update makes it sensitive to ordering: early
deaths produce short trajectories with little signal about later
segments. DQN was run as a parallel mechanistic contrast
(Section~\ref{sec:dqn_null}).

\section{Results}

\subsection{Algorithm Comparison}
\label{sec:section1}

\subsubsection{Results Summary}

\begin{figure}[htbp]
  \centering
  \includegraphics[width=0.85\textwidth]{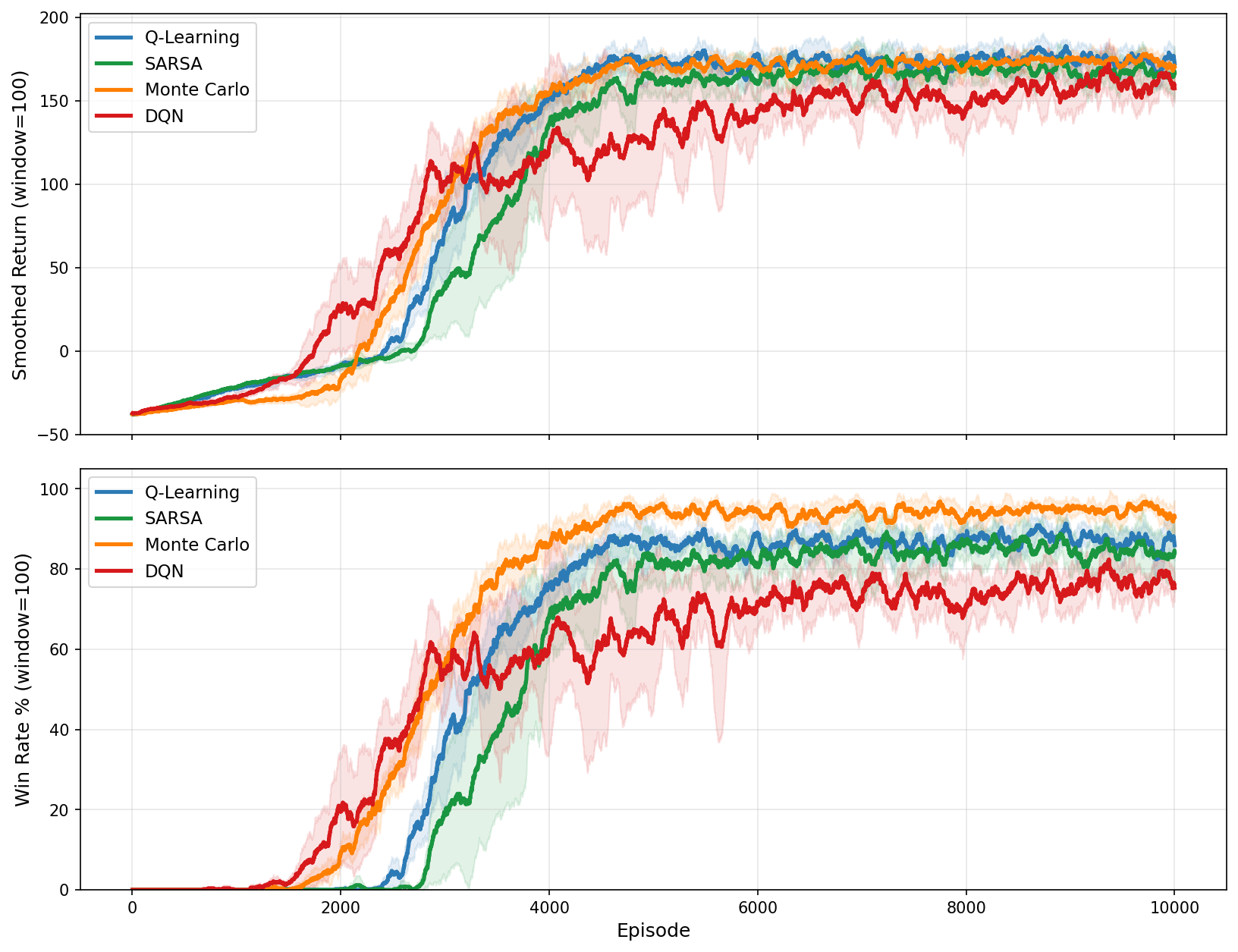}
  \caption{Learning curves on v3 (enemies). Top: smoothed average
    return (100-episode window). Bottom: smoothed win rate.
    Shaded bands show $\pm 1$ std across 5 seeds. DQN rises
    earliest but achieves the lowest final win rate and the
    greatest variance across seeds; MC achieves the highest
    final win rate.}
  \label{fig:learning_curves}
\end{figure}

\begin{figure}[htbp]
  \centering
  \includegraphics[width=0.85\textwidth]{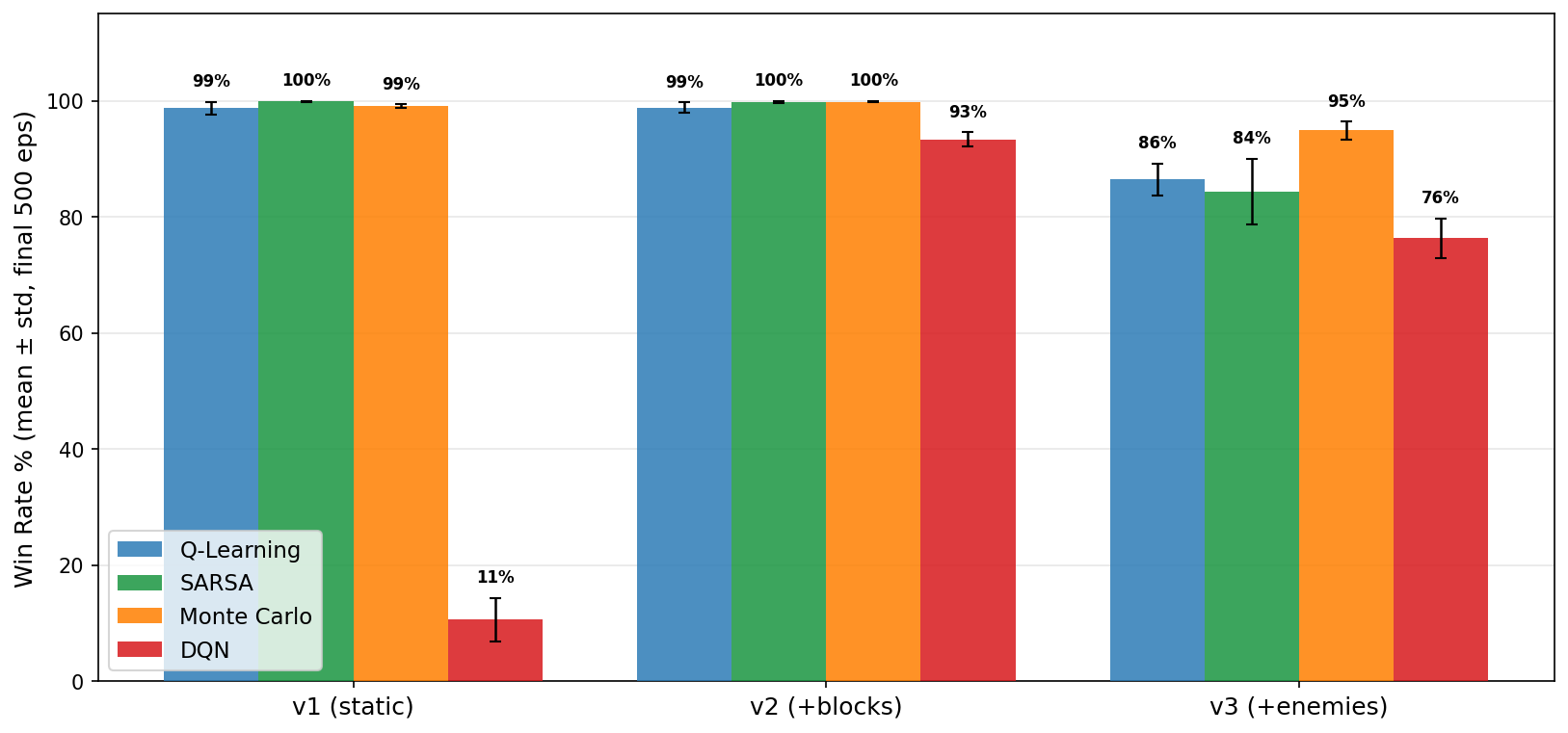}
  \caption{Final win rate across all agents and level versions
    (mean $\pm$ std, final 500 episodes). DQN's catastrophic
    failure on v1 ($10.6\%$) and recovery on v2 ($93.4\%$) highlight
    the critical role of intermediate reward density.}
  \label{fig:bar_chart}
\end{figure}

Table~\ref{tab:section1_results} reports final win rate, average
return, and convergence speed for all 12 agent-level
combinations. Figure~\ref{fig:learning_curves} shows smoothed
learning curves (return and win rate) for all four agents on v3.
Figure~\ref{fig:bar_chart} shows final win rates across all
agent-level combinations.

\begin{table}[htbp]
  \centering
  \caption{Algorithm comparison results: mean $\pm$ std across 5 seeds (42--46),
           10,000 episodes, $\text{REWARD\_RIGHT} = +0.5$,
           final 500 episodes. Bold = best value within each level group (v1, v2, v3).
           ``N/A'' = threshold not reached within 10,000 episodes;
           ``---'' = mechanic not present in that level version.}
  \label{tab:section1_results}
  \renewcommand{\arraystretch}{1.15}
  \resizebox{\textwidth}{!}{%
  \begin{tabular}{llccccccc}
    \toprule
    Agent & Level & Win Rate & Avg Return & $\to$50\% & $\to$80\%
          & Enemies & Bricks & Q-boxes \\
    \midrule
    Q-Learning & v1 & $98.8 \pm 1.1\%$ & $191.7 \pm 2.1$ & $1345$ & $2122$ & ---    & ---           & ---           \\
    Q-Learning & v2 & $98.9 \pm 0.9\%$ & $205.5 \pm 3.5$ & $1241$ & $1964$ & ---    & $0.63$        & $4.74$        \\
    Q-Learning & v3 & $86.5 \pm 2.7\%$ & $173.3 \pm 4.7$ & $3157$ & $4029$ & $3.76$ & $1.33$        & $3.96$        \\
    \midrule
    SARSA      & v1 & $\mathbf{100.0 \pm 0.1\%}$ & $\mathbf{193.9 \pm 0.3}$ & $1054$ & $1553$ & ---    & ---           & ---           \\
    SARSA      & v2 & $99.8  \pm 0.1\%$ & $\mathbf{207.6 \pm 2.5}$ & $1126$ & $1645$ & ---    & $0.25$        & $4.94$        \\
    SARSA      & v3 & $84.4  \pm 5.6\%$ & $167.2 \pm 8.5$ & $3510$ & $4446$ & $\mathbf{3.78}$ & $1.52$        & $3.48$        \\
    \midrule
    Monte Carlo & v1 & $99.1 \pm 0.3\%$ & $177.3 \pm 1.0$ & $\mathbf{996}$  & $\mathbf{1545}$ & ---    & ---           & ---           \\
    Monte Carlo & v2 & $\mathbf{99.9 \pm 0.2\%}$ & $202.1 \pm 3.6$ & $\mathbf{756}$  & $\mathbf{1173}$ & ---    & $\mathbf{2.04}$ & $\mathbf{6.78}$ \\
    Monte Carlo & v3 & $\mathbf{94.9 \pm 1.5\%}$ & $\mathbf{173.9 \pm 3.7}$ & $2816$ & $\mathbf{3493}$ & $3.28$ & $\mathbf{4.05}$ & $\mathbf{5.20}$ \\
    \midrule
    DQN        & v1 & $10.6 \pm 3.7\%$  & $13.0 \pm 12.6$ & N/A    & N/A    & ---    & ---           & ---           \\
    DQN        & v2 & $93.4 \pm 1.2\%$  & $194.3 \pm 3.3$ & $1242$ & $1899$ & ---    & $0.08$        & $4.75$        \\
    DQN        & v3 & $76.4 \pm 3.4\%$  & $159.4 \pm 8.0$ & $\mathbf{2643}$ & $4016$ & $1.38$ & $0.39$        & $4.60$        \\
    \bottomrule
  \end{tabular}%
  }
\end{table}

\subsubsection{Tabular Agents on v1 and v2}

All three tabular agents achieve near-perfect win rates on v1 and
v2 ($\geq 98.8\%$). MC converges fastest on v1 (to 80\%:
$1545 \pm 160$ episodes), followed closely by SARSA
($1553 \pm 100$) and Q-Learning ($2122 \pm 404$). The difference
between on-policy SARSA and off-policy Q-Learning on early
convergence is consistent with SARSA's tendency to take safer
paths during exploration: in a level where falling into a gap is
catastrophic, conservatism during training accelerates early
learning.

On v2, MC leads on convergence speed and win rate. MC v2
reaches 80\% win rate in $1173 \pm 98$ episodes---the fastest
of any agent on any level---and achieves the highest win rate
($99.9\% \pm 0.2\%$). MC also hits substantially more blocks:
$2.04$ bricks and $6.78$ question blocks per episode, compared
to $0.63$/$4.74$ for Q-Learning and $0.25$/$4.94$ for SARSA.

\subsubsection{v3 Performance and the MC Advantage}

Adding enemies (v3) reduces win rates across all agents, but
differentially. Q-Learning drops from 98.9\% to $86.5\% \pm
2.7\%$; SARSA drops to $84.4\% \pm 5.6\%$; MC drops only to
$\mathbf{94.9\% \pm 1.5\%}$. The Monte-Carlo agent's advantage is statistically significant against Q-Learning (Welch's t-test: p = 0.0008, d = 3.85), SARSA (Welch's t-test: p = 0.012, d = 2.56) and DQN (Welch's t-test: p = 0.0001, d = 7.04).

MC's advantage on v3 is not merely
quantitative; it is behavioral. MC v3 hits $4.05$ bricks and
$5.20$ question blocks per episode; Q-Learning hits only
$1.33$ and $3.96$, SARSA $1.52$ and $3.48$.

This reflects the fundamental difference between episode-level
and step-level updates. TD methods propagate value
\emph{locally}: the value of a state reflects expected future
return from that state, incentivizing the most direct path to
the flagpole. MC propagates value \emph{globally}: every
(state, action) pair on a winning trajectory is incremented by
the \emph{total} return of that episode, including all
intermediate rewards collected along the way. On a level where
bricks ($+1$) and question blocks ($+3$) are scattered across
every winning path, MC naturally discovers and exploits them.
All tabular agents stomp enemies at comparable rates ($3.28$--$3.78$
kills/ep), but MC collects substantially more blocks along the
way: a richer, higher-return strategy.

SARSA's higher standard deviation on v3 ($5.6\%$ vs $2.7\%$
for Q-Learning) reflects seed sensitivity to early-episode
exploration: SARSA's on-policy conservatism, which accelerates
convergence on static levels, can become a liability when enemies
introduce stochastic death conditions that vary across seeds.

\subsubsection{DQN: Catastrophic Failure on v1, Recovery on v2}

DQN's behavior is qualitatively different from all tabular
agents. On v1 (no intermediate rewards), DQN achieves only
$10.6\% \pm 3.7\%$ win rate after 10,000 episodes and
\emph{never stabilizes}; peaks and collapses occur repeatedly
across all five seeds, with return variance of $\pm 12.55$. The
cause is replay buffer imbalance: on v1, the directional reward
($+0.5$/step) accumulates only modestly in short episodes that
end in death, while the death penalty ($-50$) is large and
frequent. Death transitions flood the 50k experience buffer,
starving the policy of positive signal and destabilizing
training.

On v2, DQN recovers dramatically to $93.4\% \pm 1.2\%$. The
addition of question block rewards ($+3$) diversifies the reward
signal throughout the level, giving the replay buffer a richer
mix of positive transitions to learn from. DQN v2 averages
$4.75$ question blocks hit per episode but only $0.08$ bricks,
suggesting it learns specific landmark interactions rather than
a general strategy of hitting objects from below.

On v3, DQN achieves $76.4\% \pm 3.4\%$---the weakest of all
four agents. Enemy kills average only $1.38$/ep versus
$3.28$--$3.78$ for tabular agents. DQN generalizes a
\emph{cautious} policy: it learns to treat enemies as threats
to be avoided rather than targets worth the risk of $+5$.
This is a rational response to the generalization pressure of
the MLP: a single network must represent Mario's value across
all states, and the gradient signal from enemy collisions
($-50$) creates a strong repulsive feature that dominates the
attraction of the stomp reward.

\subsection{Curriculum Learning Experiment}
\label{sec:section2}

\subsubsection{Hypothesis and Design}

We hypothesize that World 1-1's canonical segment ordering
(A$\to$B$\to$C$\to$D$\to$E$\to$F) accelerates reinforcement learning
because it introduces mechanics progressively: an implicit
curriculum \cite{bengio2009}. To test this, we train Monte
Carlo on all 12 conditions and compare win rate, convergence
speed, learning efficiency (area under the win-rate curve,
AUC), and catastrophic failure rate.

\subsubsection{Results}

\begin{figure}[htbp]
  \centering
  \includegraphics[width=0.85\textwidth]{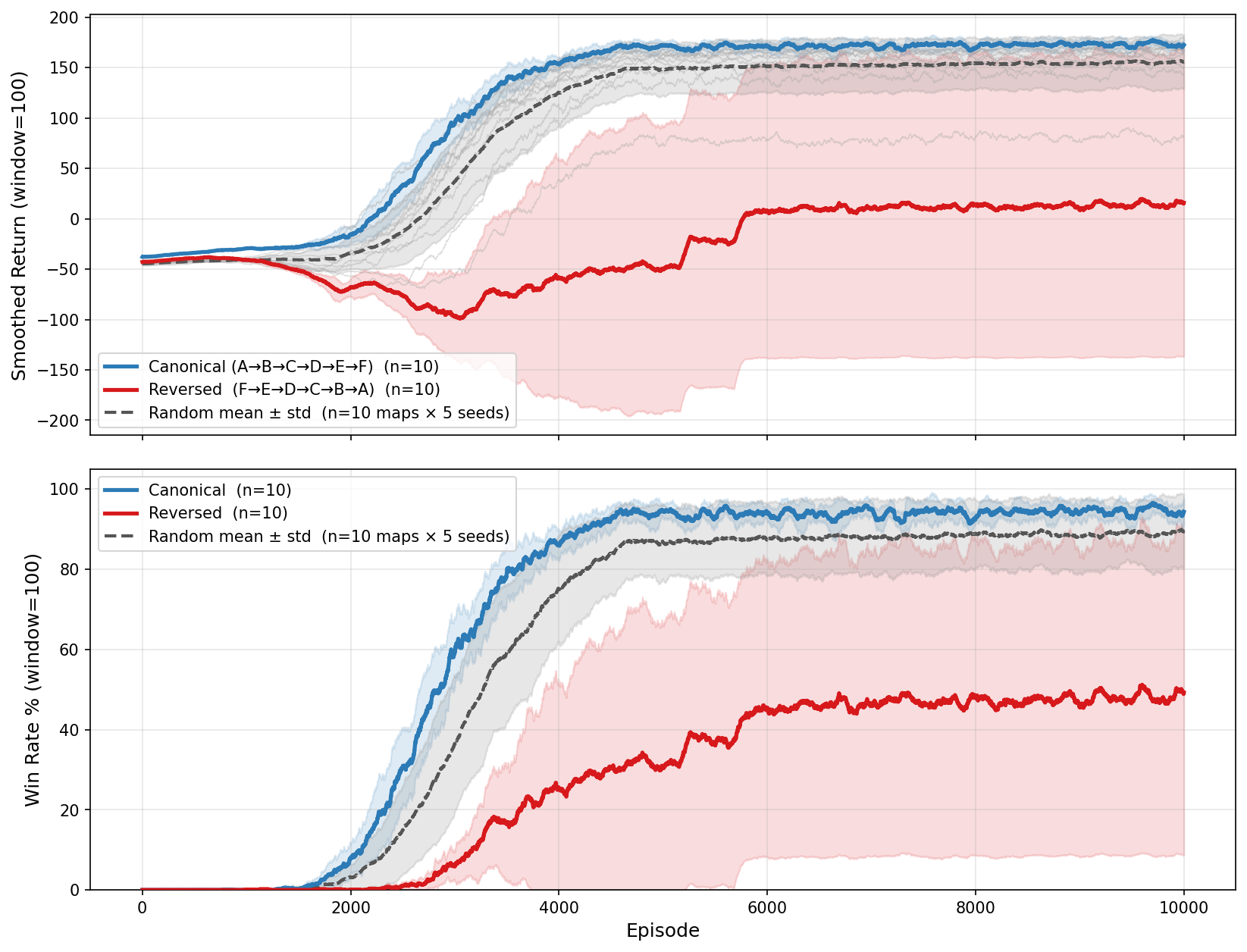}
  \caption{Curriculum experiment learning curves (Monte Carlo on v3).
    Top: smoothed average return. Bottom: smoothed win rate.
    Canonical (blue) converges fastest and most consistently.
    Reversed (red) is severely delayed and shows catastrophic
    failure variance. Dashed gray band: mean $\pm$ std of 10
    random permutations ($n = 10$ maps $\times$ 5 seeds).}
  \label{fig:curriculum_curves}
\end{figure}

\begin{figure}[htbp]
  \centering
  \includegraphics[width=0.85\textwidth]{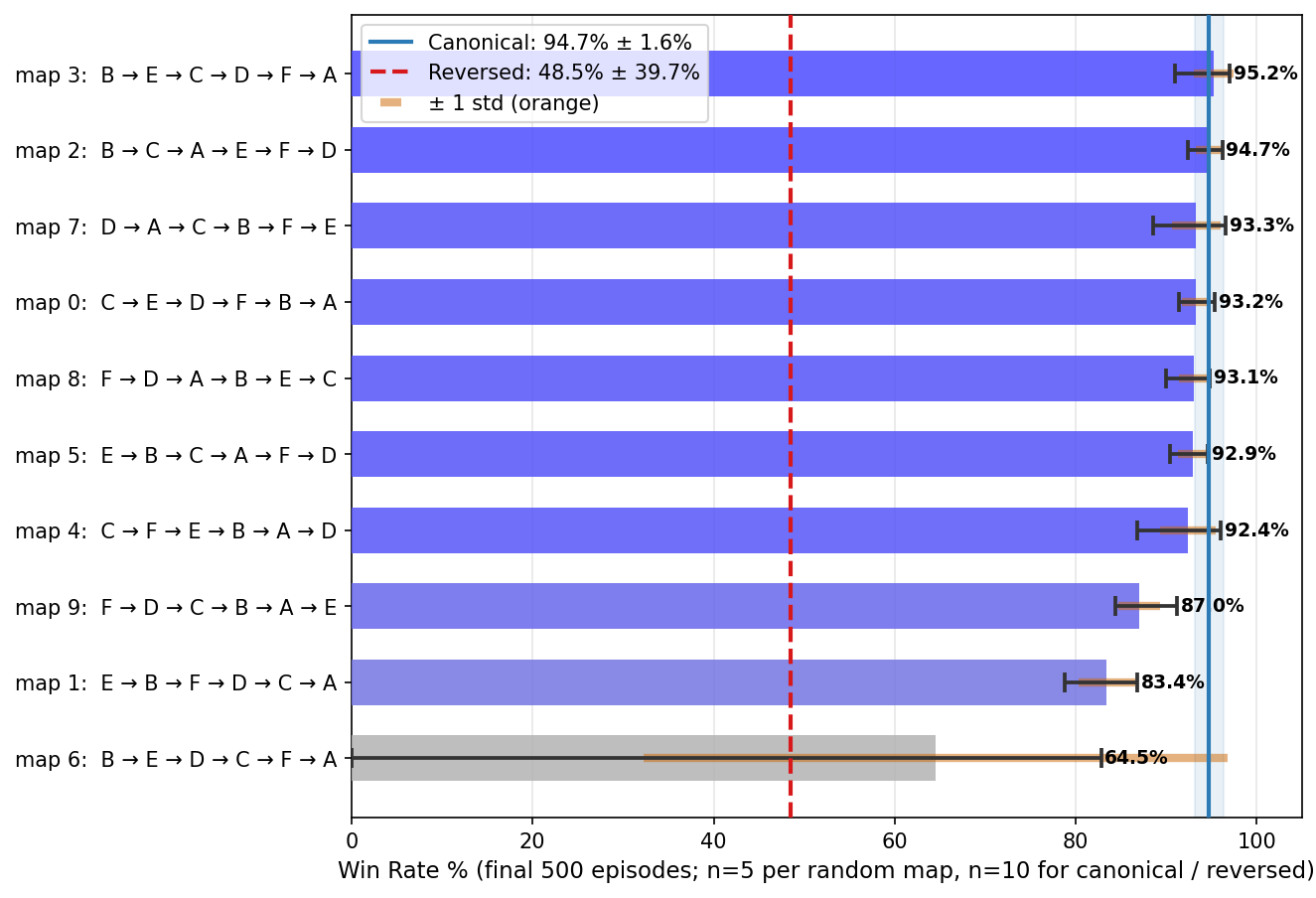}
  \caption{Per-map random win rates vs.\ canonical and reversed
    baselines (MC, mean across 5 seeds, final 500 episodes).
    Maps are sorted high to low. Map~3 ($\text{B}{\to}\text{E}{\to}\text{C}{\to}\text{D}{\to}\text{F}{\to}\text{A}$)
    is the best single random map at $95.2\%$; Map~6
    ($\text{B}{\to}\text{E}{\to}\text{D}{\to}\text{C}{\to}\text{F}{\to}\text{A}$)
    falls to $64.5\%$ with one catastrophic seed failure.}
  \label{fig:permap}
\end{figure}

Figure~\ref{fig:curriculum_curves} shows the learning curves
for canonical, reversed, and the random mean band.
Table~\ref{tab:curriculum_results} presents the full quantitative summary. Figure~\ref{fig:permap}
shows per-map win rates for all 10 random permutations.

\begin{table}[htbp]
  \centering
  \caption{Curriculum experiment quantitative summary: Monte Carlo on v3,
           mean $\pm$ std across 5 seeds (42--46) with additional seeds (54-58) for canonical and reversed, 10,000 episodes.
           Bold = best value per column. Failures = seeds with final win rate $< 10\%$.}
  \label{tab:curriculum_results}
  \renewcommand{\arraystretch}{1.2}
  \resizebox{\textwidth}{!}{%
  \begin{tabular}{lcccccc}
    \toprule
    Condition & Win Rate & Avg Return & $\to$50\% ep & $\to$80\% ep & AUC & Failures \\
    \midrule
    Canonical   & $\mathbf{94.7 \pm 1.6\%}$  & $\mathbf{173.4 \pm 3.4}$  & $\mathbf{2771 \pm 133}$   & $\mathbf{3472 \pm 231}$   & $\mathbf{67.2 \pm 2.0\%}$  & $\mathbf{0/10}$ \\
    Reversed    & $48.5 \pm 39.7\%$            & $14.8 \pm 151.8$           & $6586 \pm 2865$            & $7142 \pm 2447$            & $27.5 \pm 22.8\%$  & $4/10$          \\
    Random mean & $89.0 \pm 8.9\%$             & $155.3 \pm 26.3$           & $3333 \pm 640$             & $4065 \pm 657$             & $60.0 \pm 8.0\%$            & $1/50$         \\
    \bottomrule
  \end{tabular}%
  }
\end{table}

\subsubsection{Canonical Ordering is Fastest and Most Robust}

Canonical ordering achieves the highest final win rate
($94.7\% \pm 1.6\%$), significantly above the random pool ($89.0\% \pm 8.9\%$) and reversed condition ($94.7\% \pm 1.6\%$) as well as the fastest convergence to 50\% win rate 
($2771 \pm 133$ episodes) and the highest AUC ($67.2\% \pm
2.0\%$). Critically, it is the \emph{only} condition with zero
catastrophic failures: all ten seeds converge reliably.

The mechanism is consistent with MC's update structure.
Canonical ordering begins with Segment A (open ground, one
Goomba), giving the agent a high probability of episode
completion in early training. Early successes produce positive
Monte Carlo returns that bootstrap a stable value function
before $\epsilon$ decays to its exploration floor. Each
subsequent segment introduces a harder mechanic (pipes,
gaps, enemy clusters) after the agent has already learned
to navigate the preceding one. By the time the agent encounters
the D-segment enemy gauntlet, it has already internalized the
winning trajectory up to that point.

\subsubsection{Reversed Ordering is Severely Penalized}

Reversed ordering with n extended to 10 seeds (F$\to$E$\to$D$\to$C$\to$B$\to$A) achieves
only $48.5\% \pm 39.7\%$ win rate---a 46 percentage-point gap
from canonical---with 4 seeds failing completely (final win rate
$< 10\%$). The standard deviation of $39.7\%$ is catastrophic
for a mean this low: it indicates that any individual seed could
produce a non-functional agent. Welch's t-test confirms that the difference between canonical and reverse ordering's mean competion rate is statistically significant with $p=0.005$ and $d=1.64$.

The mechanism is the inverse of the canonical case. Segment F
contains the final staircase, two pipes, and two Goombas;
Segment D contains nine Goombas in clusters. An agent that must
navigate these as its first experience has a very low chance of
completing an episode. As $\epsilon$ decays, the policy is
trained primarily on failed short episodes, reinforcing early
death and suppressing the value of later states. Recovery is
possible (four of ten seeds do recover), but slow: the mean
convergence to 50\% win rate is $6586 \pm 2865$ episodes versus $2771 \pm 133$ for canonical. This difference is also statistically significant (Welch's t-test: $p = 0.0023$, $d = 1.88$).

\subsubsection{Random Orderings Cluster Between Canonical and Reversed}

The ten random permutations produce a mean win rate of
$89.0\% \pm 8.9\%$, intermediate between canonical and
reversed.  However, this mean conceals substantial variance:
the best single random map (Map~3:
B$\to$E$\to$C$\to$D$\to$F$\to$A) achieves $95.2\%$---statistically
equivalent to canonical on win rate (Welch's $t$-test,
$p \gg 0.05$)---while Map~6
(B$\to$E$\to$D$\to$C$\to$F$\to$A) falls to $64.5\%$ with one
catastrophic seed failure.

The difference between Map~3 and Map~6 is instructive:
both begin with Segment B (pipes, Goombas), but Map~3 follows
with C (gaps and blocks) then progresses to D (enemy gauntlet),
while Map~6 puts D immediately after B. The Enemy Gauntlet
(Segment D) appears to be the critical difficulty spike: any
ordering that places D in position 2 or 3 without a gentler
preceding segment risks early catastrophic failure.

Canonical ordering uniquely guarantees that D is encountered
fourth, after A (easy), B (moderate), and C (moderate with
gaps). No random map reliably matches all three of canonical's
criteria---highest win rate, fastest convergence, and zero
failures---simultaneously.

\section{Discussion}
\label{sec:discussion}

\subsection{Why MC Outperforms DQN on a Natively Discrete Level}

The standard argument for DQN is that it scales to continuous
or large discrete state spaces where tabular methods are
intractable. World 1-1's state space is large in theory
(column $\times$ row $\times$ jump $\times$ viewport tuples)
but narrow in practice: most states are never visited because
Mario can only occupy one of a few rows when grounded and the
jump arc is deterministic. The tabular Q-table's
\texttt{defaultdict} implementation stores only visited states,
meaning the effective Q-table is much smaller than the
theoretical state count.

Under these conditions, DQN's function approximation introduces
a penalty that tabular methods avoid: generalization across
similar but distinct states. A network that sees Mario at
column 100, row 11, near a Goomba must generalize that
experience across nearby columns and rows. But the level is
not spatially uniform; enemy positions, gaps, and block
arrangements differ column by column. Generalizing the wrong
way creates a cautious average policy rather than a
location-specific optimal one. Tabular methods memorize the
optimal action for every visited state without smoothing across
neighbors, which is exactly right for a discrete, non-uniform
environment.

\subsection{DQN's Advantage is Reward-Density Dependent}

DQN's recovery from catastrophic failure on v1 ($10.6\%$) to
strong performance on v2 ($93.4\%$) illustrates when DQN's
replay buffer is \emph{helpful}. On v1, the dominant return signals are $+100$ (rare win)
and $-50$ (common death), with only modest directional
reward ($+0.5$/step) in between. The
replay buffer skews heavily toward death transitions, and the
network cannot bootstrap because positive experiences are too
rare to maintain. On v2, question blocks ($+3$) are distributed
across the level and encountered frequently during random
exploration, providing a diverse return signal that stabilizes
the replay buffer.

This frames DQN as the appropriate choice for problems where
the state space is too large for tabular methods
\emph{and} where reward density is high enough to provide
diverse signal to the replay buffer. World 1-1 under
directional reward does not satisfy the first condition:
the effective state space is narrow and tabular methods
handle it exactly. DQN's experience replay provides no
final-performance advantage over TD methods under these conditions.

\subsection{The DQN Null Curriculum Effect}
\label{sec:dqn_null}

\begin{figure}[htbp]
  \centering
  \includegraphics[width=0.95\textwidth]{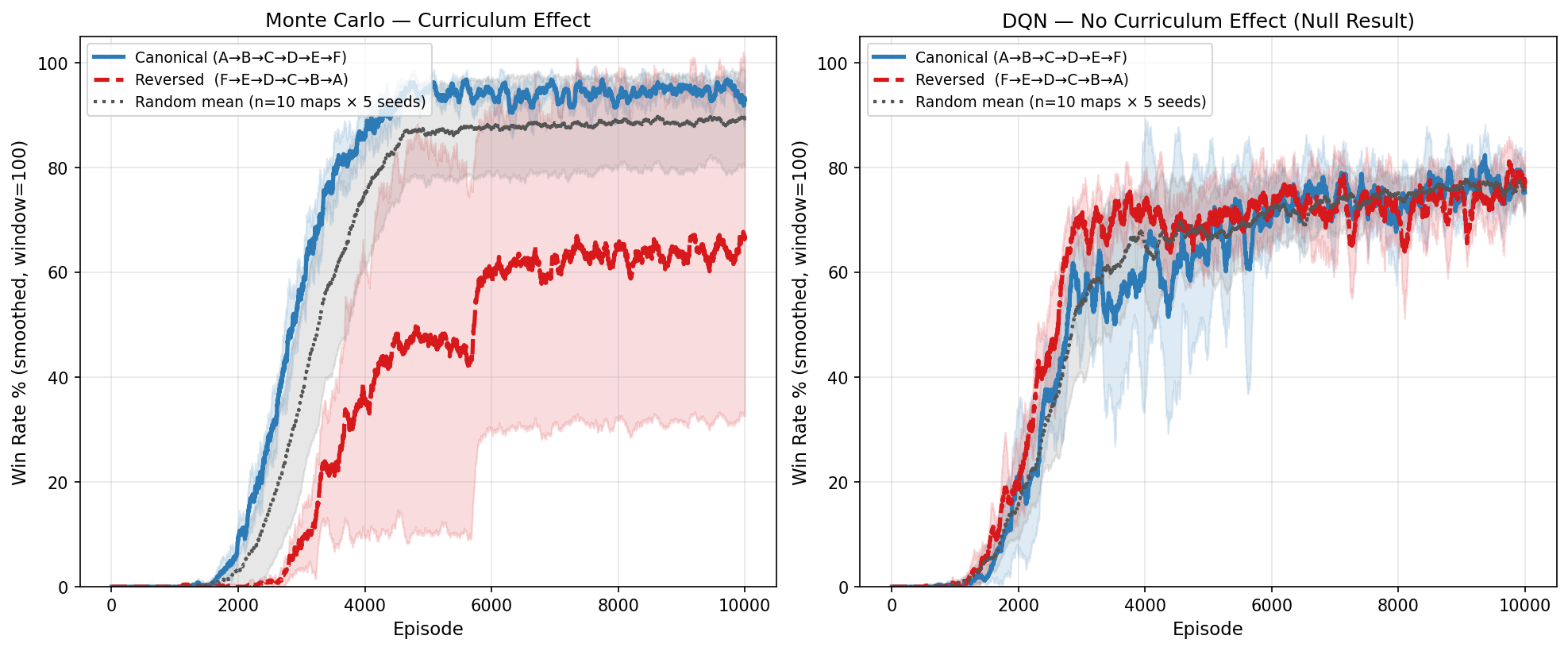}
  \caption{Curriculum ordering effect: MC (left, 5 seeds, sensitive) vs.\
    DQN (right, 5 seeds, insensitive). MC canonical and reversed diverge
    dramatically; all three DQN conditions overlap completely.}
  \label{fig:dqn_contrast}
\end{figure}

To isolate the mechanism behind MC's curriculum sensitivity,
we ran the same 12 conditions with DQN (Figure~\ref{fig:dqn_contrast},
Table~\ref{tab:dqn_curriculum}). DQN is completely insensitive
to segment ordering: canonical, reversed, and random conditions
all converge to ${\approx}76\%$ win rate. 

A one-way ANOVA across canonical, reversed, and random conditions found no significant effect of segment ordering on DQN's final win rate ($F(2, 57) = 0.19$, $p = 0.82$, $\eta^2 = 0.007$). Pairwise Welch t-tests confirmed this: canonical vs. reversed ($p = 0.82$, $d = -0.15$) and canonical vs. random pool ($p = 0.90$, $d = 0.08$) both showed effects below $d = 0.2$, confirming that the differences between these distributions were not statistically significant. 

\begin{table}[htbp]
  \centering
  \caption{DQN curriculum results (null result): mean $\pm$ std
           across 5 seeds.}
  \label{tab:dqn_curriculum}
  \renewcommand{\arraystretch}{1.15}
  \begin{tabular}{lccc}
    \toprule
    Condition & Win Rate & AUC & Failures \\
    \midrule
    Canonical   & $76.4\% \pm 3.4\%$ & $52.8\% \pm 4.0\%$ & 0/5  \\
    Reversed    & $77.0\% \pm 4.6\%$ & $55.9\% \pm 2.5\%$ & 0/5  \\
    Random mean & $76.2\% \pm 2.5\%$ & $53.5\% \pm 4.2\%$ & 0/50 \\
    \bottomrule
  \end{tabular}
\end{table}

The mechanism is DQN's experience replay buffer. At every
gradient update, the network samples uniformly from the full
history of 50,000 transitions, regardless of the order in which
they were collected. The temporal signal created by segment
ordering---experiencing easy segments before hard ones---is
completely erased by this uniform sampling. From the network's
perspective, a transition collected in episode 1 (easy segment)
is statistically indistinguishable from one collected in
episode 5000 (hard segment).

The replay buffer's insensitivity to ordering and its ability
to decorrelate training samples are two sides of the same
architectural coin. Our results suggest that future
curriculum learning research should distinguish between
update-mechanism-sensitive agents (MC, and likely other
episode-level learners) and update-mechanism-robust agents
(DQN, and other replay-based methods).

An open question is where SARSA and Q-Learning fall on this
spectrum. Both are step-level TD methods without a replay
buffer, but they update on single transitions rather than full
episodes. They may show intermediate curriculum sensitivity:
harder early segments delay convergence but do not induce the
catastrophic collapse that MC's episode-level mechanism can
produce. We leave this comparison to future work.

\subsection{Reward Density as a First-Class Variable in Method Selection}

Practitioners selecting between tabular methods and DQN
typically focus on state space size as the deciding variable.
Our results suggest reward density deserves equal
consideration. On v1, DQN fails not because the state space
is too large but because the replay buffer is flooded with
death penalties, starving the network of positive signal. On
v2, sufficient intermediate reward density stabilizes training
and DQN recovers to $93.4\%$. State space structure and reward
density are jointly necessary conditions --- neither alone
predicts whether DQN will succeed. Future work applying DQN
to similarly bounded discrete environments should treat reward
density as a first-class design variable.

\section{Conclusion}

We compared four reinforcement learning algorithms on a custom
implementation of Super Mario Bros World 1-1 across three levels
of increasing complexity. Our main findings are:

\begin{enumerate}
  \item Tabular agents converge on all three levels; DQN
        converges on v2 and v3 but fails catastrophically on v1
        ($10.6\% \pm 3.7\%$). Among converging agents, Monte
        Carlo achieves the highest win rate on the full level
        with enemies ($94.7\% \pm 1.6\%$) through an emergent
        strategy of collecting more intermediate rewards en route
        to the flagpole. DQN achieves the lowest ($76.4\% \pm
        3.4\%$) due to overgeneralized enemy avoidance.

  \item DQN's performance is reward-density dependent. On v1,
        the death penalty ($-50$) dominates the replay buffer
        and the policy never stabilizes. On v2, question block
        rewards provide enough signal diversity to stabilize
        training and DQN recovers to $93.4\% \pm 1.2\%$. Under
        this reward structure, tabular methods achieve higher final
        win rates and average returns on v3, though DQN reaches
        50\% win rate earliest---consistent with its early
        generalization advantage before the ceiling of its cautious
        policy becomes apparent.

  \item The curriculum experiment strongly supports the
        hypothesis that World 1-1's design is pedagogically
        structured. Canonical ordering is the fastest, most
        efficient, and most robust condition, producing zero
        catastrophic failures in 10/10 seeds. Reversed ordering
        fails four seeds entirely. No random permutation matches
        all three of canonical's criteria simultaneously.

  \item The curriculum effect is driven by the update
        mechanism. DQN shows a complete null curriculum effect
        because experience replay erases temporal ordering
        information. This implicates the MC episode-level
        update, which requires a full trajectory to generate a
        return signal, as the driver of ordering sensitivity.
\end{enumerate}

Future work should extend the curriculum comparison to SARSA
and Q-Learning to determine whether TD step-level updates
produce intermediate ordering sensitivity, and should probe the
crossover complexity at which DQN definitively surpasses tabular
methods.


\end{document}